\documentclass{article}

\usepackage{microtype}
\usepackage[dvipdfmx]{graphicx}
\usepackage{subfigure}
\usepackage{booktabs} % for professional tables

\usepackage{hyperref}

\usepackage[accepted]{synsml2023}

\usepackage{amsmath}
\usepackage{amssymb}
\usepackage{mathtools}
\usepackage{amsthm}
\usepackage{color}

\usepackage[capitalize,noabbrev]{cleveref}

\theoremstyle{plain}

\theoremstyle{definition}

\theoremstyle{remark}

\usepackage[textsize=tiny]{todonotes}
\usepackage{placeins}

\synsmltitlerunning{Generating observation guided ensembles for data assimilation with denoising diffusion probabilistic model}

\newcommand{\eqnref}[1]{\ref{eq:#1}}
\newcommand{\figref}[1]{\ref{fig:#1}}

\newcommand{\vect}[1]{\mathbf{#1}}

\newcommand{\redtext}[1]{\textcolor{black}{#1}}

\RequirePackage[normalem]{ulem} %DIF PREAMBLE
\RequirePackage{color}\definecolor{RED}{rgb}{1,0,0}\definecolor{BLUE}{rgb}{0,0,1} %DIF PREAMBLE
\providecommand{\DIFaddbegin}{} %DIF PREAMBLE
\providecommand{\DIFaddend}{} %DIF PREAMBLE
\providecommand{\DIFdelbegin}{} %DIF PREAMBLE
\providecommand{\DIFdelend}{} %DIF PREAMBLE
\providecommand{\DIFaddbeginFL}{} %DIF PREAMBLE
\providecommand{\DIFaddendFL}{} %DIF PREAMBLE
\providecommand{\DIFdelbeginFL}{} %DIF PREAMBLE
\providecommand{\DIFdelendFL}{} %DIF PREAMBLE
\newcommand{\DIFscaledelfig}{0.5}
\RequirePackage{settobox} %DIF PREAMBLE
\RequirePackage{letltxmacro} %DIF PREAMBLE
\newsavebox{\DIFdelgraphicsbox} %DIF PREAMBLE
\newlength{\DIFdelgraphicswidth} %DIF PREAMBLE
\newlength{\DIFdelgraphicsheight} %DIF PREAMBLE
\LetLtxMacro{\DIFOincludegraphics}{\includegraphics} %DIF PREAMBLE
\newcommand{\DIFaddincludegraphics}[2][]{{\color{blue}\fbox{\DIFOincludegraphics[#1]{#2}}}} %DIF PREAMBLE
\newcommand{\DIFdelincludegraphics}[2][]{% %DIF PREAMBLE
\sbox{\DIFdelgraphicsbox}{\DIFOincludegraphics[#1]{#2}}% %DIF PREAMBLE
\settoboxwidth{\DIFdelgraphicswidth}{\DIFdelgraphicsbox} %DIF PREAMBLE
\settoboxtotalheight{\DIFdelgraphicsheight}{\DIFdelgraphicsbox} %DIF PREAMBLE
\scalebox{\DIFscaledelfig}{% %DIF PREAMBLE
\parbox[b]{\DIFdelgraphicswidth}{\usebox{\DIFdelgraphicsbox}\\[-\baselineskip] \rule{\DIFdelgraphicswidth}{0em}}\llap{\resizebox{\DIFdelgraphicswidth}{\DIFdelgraphicsheight}{% %DIF PREAMBLE
\setlength{\unitlength}{\DIFdelgraphicswidth}% %DIF PREAMBLE
\begin{picture}(1,1)% %DIF PREAMBLE
\thicklines\linethickness{2pt} %DIF PREAMBLE
{\color[rgb]{1,0,0}\put(0,0){\framebox(1,1){}}}% %DIF PREAMBLE
{\color[rgb]{1,0,0}\put(0,0){\line( 1,1){1}}}% %DIF PREAMBLE
{\color[rgb]{1,0,0}\put(0,1){\line(1,-1){1}}}% %DIF PREAMBLE
\end{picture}% %DIF PREAMBLE
}\hspace*{3pt}}} %DIF PREAMBLE
} %DIF PREAMBLE
\LetLtxMacro{\DIFOaddbegin}{\DIFaddbegin} %DIF PREAMBLE
\LetLtxMacro{\DIFOaddend}{\DIFaddend} %DIF PREAMBLE
\LetLtxMacro{\DIFOdelbegin}{\DIFdelbegin} %DIF PREAMBLE
\LetLtxMacro{\DIFOdelend}{\DIFdelend} %DIF PREAMBLE
\DeclareRobustCommand{\DIFaddbegin}{\DIFOaddbegin \let\includegraphics\DIFaddincludegraphics} %DIF PREAMBLE
\DeclareRobustCommand{\DIFaddend}{\DIFOaddend \let\includegraphics\DIFOincludegraphics} %DIF PREAMBLE
\DeclareRobustCommand{\DIFdelbegin}{\DIFOdelbegin \let\includegraphics\DIFdelincludegraphics} %DIF PREAMBLE
\DeclareRobustCommand{\DIFdelend}{\DIFOaddend \let\includegraphics\DIFOincludegraphics} %DIF PREAMBLE
\LetLtxMacro{\DIFOaddbeginFL}{\DIFaddbeginFL} %DIF PREAMBLE
\LetLtxMacro{\DIFOaddendFL}{\DIFaddendFL} %DIF PREAMBLE
\LetLtxMacro{\DIFOdelbeginFL}{\DIFdelbeginFL} %DIF PREAMBLE
\LetLtxMacro{\DIFOdelendFL}{\DIFdelendFL} %DIF PREAMBLE
\DeclareRobustCommand{\DIFaddbeginFL}{\DIFOaddbeginFL \let\includegraphics\DIFaddincludegraphics} %DIF PREAMBLE
\DeclareRobustCommand{\DIFaddendFL}{\DIFOaddendFL \let\includegraphics\DIFOincludegraphics} %DIF PREAMBLE
\DeclareRobustCommand{\DIFdelbeginFL}{\DIFOdelbeginFL \let\includegraphics\DIFdelincludegraphics} %DIF PREAMBLE
\DeclareRobustCommand{\DIFdelendFL}{\DIFOaddendFL \let\includegraphics\DIFOincludegraphics} %DIF PREAMBLE
\RequirePackage{listings} %DIF PREAMBLE
\RequirePackage{color} %DIF PREAMBLE
\lstdefinelanguage{DIFcode}{ %DIF PREAMBLE
  moredelim=[il][\color{red}\sout]{\%DIF\ <\ }, %DIF PREAMBLE
  moredelim=[il][\color{blue}\uwave]{\%DIF\ >\ } %DIF PREAMBLE
} %DIF PREAMBLE
\lstdefinestyle{DIFverbatimstyle}{ %DIF PREAMBLE
	language=DIFcode, %DIF PREAMBLE
	basicstyle=\ttfamily, %DIF PREAMBLE
	columns=fullflexible, %DIF PREAMBLE
	keepspaces=true %DIF PREAMBLE
} %DIF PREAMBLE
\lstnewenvironment{DIFverbatim}{\lstset{style=DIFverbatimstyle}}{} %DIF PREAMBLE
\lstnewenvironment{DIFverbatim*}{\lstset{style=DIFverbatimstyle,showspaces=true}}{} %DIF PREAMBLE
\begin{document}

\twocolumn[
\synsmltitle{Generating observation guided ensembles for data assimilation with denoising diffusion probabilistic model}
%\synsmltitle{Generative Ensemble Kalman Filter with denosing diffusion probabilistic model}
%\synsmltitle{Ensemble Free Data Assimilation method with probabilistic denosing diffusion model}

% It is OKAY to include author information, even for blind
% submissions: the style file will automatically remove it for you
% unless you've provided the [accepted] option to the synsml2023
% package.

% List of affiliations: The first argument should be a (short)
% identifier you will use later to specify author affiliations
% Academic affiliations should list Department, University, City, Region, Country
% Industry affiliations should list Company, City, Region, Country

% You can specify symbols, otherwise they are numbered in order.
% Ideally, you should not use this facility. Affiliations will be numbered
% in order of appearance and this is the preferred way.
\synsmlsetsymbol{equal}{*}

\begin{synsmlauthorlist}
\synsmlauthor{Yuuichi Asahi}{jaea}
\synsmlauthor{Yuta Hasegawa}{jaea}
\synsmlauthor{Naoyuki Onodera}{jaea}
\synsmlauthor{Takashi Shimokawabe}{comp}
\synsmlauthor{Hayato Shiba}{sch}
\synsmlauthor{Yasuhiro Idomura}{jaea}
%\synsmlauthor{}{sch}
%\synsmlauthor{}{sch}
\end{synsmlauthorlist}

\synsmlaffiliation{jaea}{Computational Science and E-Systems, Japan Atomic Energy Agency, Chiba 277-0827, Japan}
\synsmlaffiliation{comp}{Information Technology Center, The University of Tokyo, Chiba 277-0882, Japan}
\synsmlaffiliation{sch}{Graduate School of Information Science, University of Hyogo, Hyogo, Japan}

\synsmlcorrespondingauthor{Yuuichi Asahi}{asahi.yuichi@jaea.go.jp}
%\synsmlcorrespondingauthor{Firstname2 Lastname2}{first2.last2@www.uk}

% You may provide any keywords that you
% find helpful for describing your paper; these are used to populate
% the "keywords" metadata in the PDF but will not be shown in the document
\synsmlkeywords{Machine Learning}

\vskip 0.3in
]

% this must go after the closing bracket ] following \twocolumn[ ...

% This command actually creates the footnote in the first column
% listing the affiliations and the copyright notice.
% The command takes one argument, which is text to display at the start of the footnote.
% The \synsmlEqualContribution command is standard text for equal contribution.
% Remove it (just {}) if you do not need this facility.

%\printAffiliationsAndNotice{}  % leave blank if no need to mention equal contribution
\printAffiliationsAndNotice{\synsmlEqualContribution} % otherwise use the standard text.

\begin{abstract}
This paper presents an ensemble data assimilation method using the pseudo ensembles generated by 
denoising diffusion probabilistic model. Since the model is trained against noisy and sparse observation data, this model can produce divergent ensembles close to observations. Thanks to the variance in generated ensembles, our proposed method displays better performance than the well-established ensemble data assimilation method when the simulation model is imperfect.
\end{abstract}

\section{Introduction}
\label{sec:introduction}

%\redtext{Numerical simulations play an important role in forecasting a real world system, such as physical, chemical and astronomical systems. In many applications, however, the states of the system cannot be measured directly.}

\redtext{Forecasting a real world system is often achieved by combining a scientific model for the time evolution of the system and an estimate of the current state of the system. In many cases, the time evolution of the system is expressed with a sort of partial differential equations and is solved numerically (i.e., numerical simulations). The current state of the system is estimated by observations or measurements. Unfortunately, neither numerical simulation models nor observations are perfect. This is where Data Assimilation (DA) plays a role to obtain a better estimate of the system using the the current and past observations together with the numerical model. Basically, DA blends the simulation states and observation data iteratively to give a reasonable estimate of the current state. For example, almost all numerical weather predictions (NWP) integrate DA methods to their simulations in order to improve the accuracy of state estimates. In weather predictions, the direct observations of the system is not feasible and the numerical model does not fully represent the time evolution of chaotic systems.}

%Data assimilation (DA) is essential to model accurately the real world with numerical simulations. Basically, DA blends the simulation states and observation data, since neither simulation models nor observations are perfect. To overcome these difficulties, for example, almost all the numerical weather predictions (NWP) integrate DA methods to their simulations in order to keep the simulated results consistent with the observed data.

%Even if the physical models in simulations are reasonable, they cannot give perfect predictions due to the chaotic nature of the world. In addition, no simulation models are perfect. 
%To overcome these difficulties, for example, almost all the numerical weather predictions (NWP) integrate DA methods to their simulations in order to keep the simulated results consistent with the observed data.

There are mainly two types of DA methods: variational methods (e.g. 3D or 4D variational method \cite{BannisterRMTS2017}) and statistical methods (e.g. the ensemble Kalman filter (EnKF) \cite{EvensenJGR1994, EvensenOcean2003}). These methods have a tolerance for incompleteness of observations (e.g. observation noises and/or unobservable states). In practice, EnKF or its derivatives are more frequently used due to their simplicity and consistency. \redtext{In EnKF, ensemble simulations (multiple simulations from different initial conditions) are performed to approximate the covariance matrix in the Kalman filter using ensemble data (See Appendix for mathematical details).} In this work, we focus on this type of approach hereinafter.

Though widely used, there are a few issues of applying EnKF to realistic models. Firstly, the so-called filter divergence can happen where the filter becomes overconfident around an incorrect state and thus the subsequent observations are ignored \cite{SacherMWR2008-1, SacherMWR2008-2, NgTellus2011}. This occurs when the variance in ensemble members is too small or cross covariance terms are too large in the ensemble covariance. For example, the limited number of ensembles, spurious long-distance covariances or the presence of model errors of different types can result in the filter divergence. Localization helps to suppress the spurious long-distance covariances, which is introduced in local ensemble transform Kalman filter (LETKF) \cite{HuntPhysicaD2007}, for example. Secondly, a large number of ensemble simulations require an extreme-scale computational power \cite{MiyoshiGRL2014}. Considering the recent extraordinary advances in neural network (NN) studies, it may be a natural consequence to use NNs to address these issues. 

%Some of these issues have been addressed with neural networks (NNs) \cite{GeerRSJ2021} in order to improve the DA accuracy \cite{OualaRS2018, TSUYUKITadashiJMSJ2022} or reduce the computational costs of these DA methods \cite{CHATTOPADHYAYJCP2023, TomizawaGMD2021, GroomsRMETS2021, PeyronRMETS2021, LiuEngana2022, BarthelemyOceanDynamics2022, yasudaArxiv2022}.

%Firstly, a large number of ensemble simulations are necessary  

%. Secondly, the presence of model errors of different types can also lead 

Indeed, NNs have been applied in order to improve the DA accuracy \cite{OualaRS2018, TSUYUKITadashiJMSJ2022} or reduce the computational costs of these DA methods \cite{CHATTOPADHYAYJCP2023, TomizawaGMD2021, GroomsRMETS2021, PeyronRMETS2021, LiuEngana2022, BarthelemyOceanDynamics2022, yasudaArxiv2022}. 

Ouala et al. demonstrated the spatio temporal interpolation by NN-based Kalman filter \yrcite{OualaRS2018}. They reported significant improvements in terms of reconstruction performance compared with conventional interpolation schemes. Tsuyuki et al. developed nonlinear DA by coupling a deep learning model and Ensemble Kalman Filter \yrcite{TSUYUKITadashiJMSJ2022}. They showed that their model outperforms EnKF in strongly nonlinear regimes despite the use of a small number of ensembles.
%Spatio temporal interpolation by NN. They show 
%neural-network (NN) representations for the time propagation of
%a Gaussian approximation of the distribution of the physical variable x. 

There are several attempts to reduce the computational costs of time evolution by approximating the simulators with NNs. Chattopadhyay et al. proposed hybrid ensemble Kalman filter (H-EnKF), which generates and evolves a large data-driven ensembles of the states of a dynamical model \yrcite{CHATTOPADHYAYJCP2023}. For quasi-geostrophic flows, their method can improve the estimation of the background error covariance matrix against EnKF by using a large amount of data-driven ensembles. Tomizawa et al. developed a reservoir computing (RC) model trained on the LETKF analysis data which outperforms LETKF when the model bias is large \yrcite{TomizawaGMD2021}. By combining LETKF and RC, their model can successfully predict the Lorenz 96 system \cite{Lorenz96} using noisy and sparse observations.

Instead of surrogate physical models, generative models are used to produce ensemble members based on variational auto-encoders (VAEs) \cite{GroomsRMETS2021, YANGJCP2021}. Other studies intend to reduce the DA cost by applying the DA method in low-dimensional latent spaces \cite{PeyronRMETS2021, LiuEngana2022}. 
%In addition, their method can predict better than LETKF when the model bias is large. 

Some recent works have focused on the Super-resolution (SR) techniques and coupled them with DA \cite{BarthelemyOceanDynamics2022, yasudaArxiv2022}. Barth{\'e}l{\'e}my et al. have proposed Super-resolution data assimilation (SRDA), which applies a statistical DA method like EnKF to the super-resolved flow fields from low resolution simulations \yrcite{BarthelemyOceanDynamics2022}. Yasuda et al. have proposed four-dimensional super-resolution data assimilation (4D-SRDA) \yrcite{yasudaArxiv2022}. In order to deal with the potential domain shift between training phase and SRDA-simulation phase, they apply SR-mixup method for domain generalization. They have demonstrated that the combination of 4D-SRDA and SR-mixup gives the robust inference particularly when observation points are spatially sparse.

Although there are many successful DA methods based on NNs, many of them focus on the cases where the model bias is absent. In practice, however, model errors or biases are inevitable and likely to be unknown.
NNs trained on a \redtext{certain model} may also suffer from risks of overfitting to \redtext{that} model. \redtext{If the model is imperfect or biased, NNs can only produce imperfect or biased results.}
%and may not work well \redtext{when that model is biased}.

%the perfect model may also suffer from risks of overfitting to the perfect model and may not work well with \bluetext{an} imperfect model. 

In this paper, we focus on a DA method which is robust against model biases. For this purpose, we propose a DA method based on the pseudo ensembles generated by the denoising diffusion probabilistic model (DDPM) \cite{HoNeurIPS2020, HoNeurIPS2021}. In our method, we perform a \redtext{non-ensemble} simulation and generate ensembles with DDPM. Generated ensembles are fed into LETKF framework as if they are coming from ensemble simulations. We then update the simulation state by \redtext{the average of ensemble mean from LETKF and the current simulation state. Hence, the information of past observation is carried only though the simulation state}. For training, we construct the dataset consisting of snapshots from numerical simulations and incomplete mock observations (both noisy and sparse observations). We train an observation-guided diffusion model to generate simulation results close to incomplete observation data. \redtext{In conventional EnKF, the numerical model is used to forecast the next states and the current states of the model is altered to fit with the observations (i.e., a Bayesian approach). In contrast, we use the numerical model to create the dataset whose distributions are learned by DDPM. Since observations are more directly used as guidance of DDPM, we can generate ensemble forecasts close to the current observation.} Thanks to the preferable feature of DDPM, the generated pseudo ensembles have diversity (or variance) while keeping similar values with observations. By using these pseudo ensembles to perform LETKF, we can update the simulation state to be \redtext{close to} observation data even if the model is biased. In addition, we do not need to perform costly ensemble simulations. Our model and datasets are available online \cite{GenerativeEnKF}.
The main contributions of this work are as follows:
\begin{itemize}
\item We develop a diffusion model that generates pseudo ensembles close to noisy and sparse observation data.
\item We demonstrate that our method is ensemble-free and robust against the imperfectness of models (model biases) and observations (noises and sparsity).
\end{itemize}

\section{Dataset and Model}
\label{sec:dataset_and_model}
\begin{figure*}[htbp]
\vskip 0.2in
\begin{center}
\includegraphics[width=1.0\hsize]{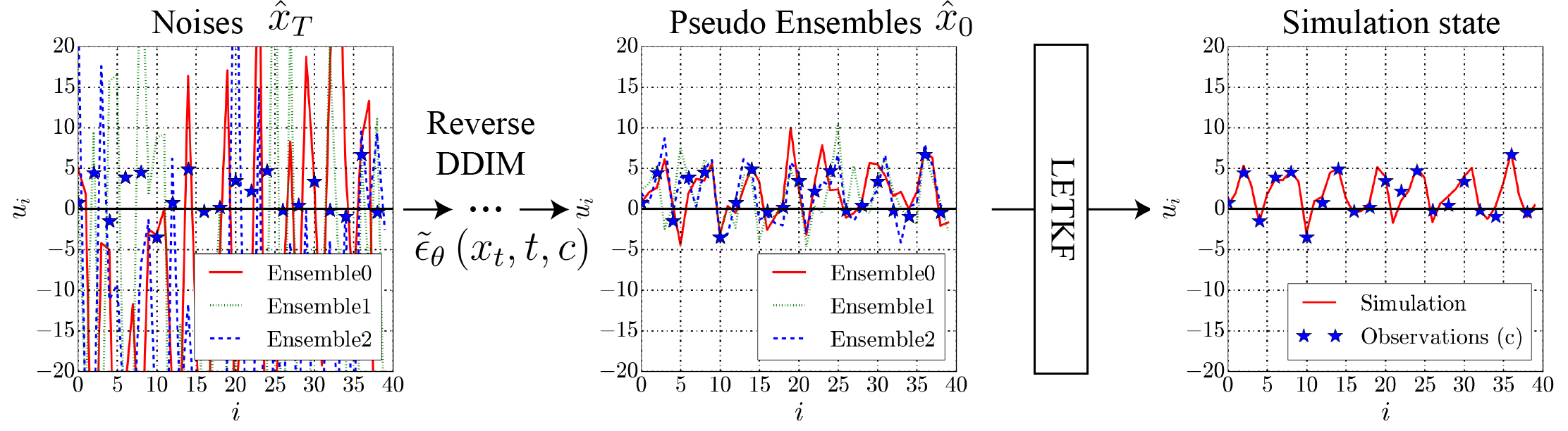}
\caption{Proposed DA method with 3 pseudo ensembles. Pseudo ensembles are generated by reverse DDIM from noises. We compute the analysis state vectors by applying LETKF to these ensembles. Then, we update the simulation state by the average of the ensemble mean and the current simulation state.}
\label{fig:architecture}
\end{center}
\vskip -0.2in
\end{figure*}

\subsection{Lorenz96 model}
\label{eq:lorenz96_model}
We employ the Lorenz96 model \cite{Lorenz96} as a minimal chaotic system defined by
\begin{eqnarray}
\frac{du_{i}}{dt}=\left(u_{i+1}-u_{i-2}\right)u_{i-1}-u_{i}+F, ~\left(\forall{i} = 1, \ldots, N\right) \label{eq:lorenz96}
\end{eqnarray}
where $F$ means the forcing parameter and subscript $i$ denotes the $i$th state variable. We employ the periodic boundary condition with $u_{-1} = u_{N-1}$, $u_{0} = u_{N}$ and $u_{N+1} = u_{1}$. This model shows chaotic behaviour with $F = 8$, but shows periodic or more chaotic behaviours with $F = 5$ and $F = 11$ (see Figs. \figref{spatio_temporal} (a) and \figref{spatio_temporal} (c)).

%\begin{eqnarray}
%\frac{dx_{i}}{dt}=\left(x_{i+1}-x_{i-2}\right)x_{i-1}-x_{i}+F, %\left(\forall{i} = 1, \ldots, N\right) \label{eq:lorenz96}
%\end{eqnarray}

In this work, we perform an observing system simulation experiment (OSSE) using the Lorenz96 model as for a proof of concept. We first conduct a raw simulation of Lorenz96 in eq. (\eqnref{lorenz96}) which is regarded as ``Nature'' run. We then add a Gaussian noise ${\cal N} \left(0, e \right)$ to the simulation results of ``Nature'' run to get mock ``Observation'' data. In this work, we choose the observation error as $e = 1$. Our primary task is to predict the ``Nature'' run from the mock ``Observation'' data.

\subsection{Lorenz96 dataset}
\label{eq:dataset}
Instead of ensemble simulations, we generate pseudo ensembles using DDPM guided by sparse and noisy ``Observation'' data. As a training dataset for DDPM, we performed the 100 Lorenz96 simulations with 1000 timesteps. \redtext{Each member} of dataset contains a snapshot of 40 state variables and mock ``Observation'' data corresponding to them. For pre-processing, we normalize each channel
to the range of $[0, 1]$.
%\begin{eqnarray}
%\frac{dx_{i}}{dt}=\left(x_{i+1}-x_{i-2}\right)x_{i-1}-x_{i}+F\left(\forall{i}=1,\ldots,N\right)) 
%\end{eqnarray}

\subsection{Conditional Diffusion Model}
\label{eq:model}
We employ the diffusion model with classifier-free guidance \cite{HoNeurIPS2021}. This model consists of forward and backward diffusion processes. The forward process uses a Markov chain where the Gaussian noise is gradually added to the original data $x_0$. For the length $T$ diffusion steps, the latent variable $x_t$ at time $t \in \left[1, \ldots, T \right]$ is 
\begin{eqnarray}
x_t = \sqrt{\overline{\alpha}_t} x_0 + \sqrt{1-\overline{\alpha}_t} \epsilon_t, \label{eq:forward}
\end{eqnarray}
where $\epsilon_t \sim {\cal N} \left(0, \mathbf{I} \right)$, $\overline{\alpha}_t = \prod_{s=1}^{t} \left( 1 - \beta_s \right)$ and $\beta_t$ is a fixed variance schedule.
The backward process or sampling process in turn generates a clean data from noises by 
\begin{eqnarray}
x_{t-1} = \frac{1}{\sqrt{1-\beta_t}} \left(x_t - \frac{\beta_t}{\sqrt{1-\alpha_t}} \tilde{\epsilon}_\theta \left(x_t, t, c \right) \right) + \sigma_t \epsilon_t, \label{eq:backward}
\end{eqnarray}
where $c$ is the conditioning information and $\sigma_t$ is the variance of the noise.
$\tilde{\epsilon}_\theta$ is the linear combination of the conditional and unconditional score estimates as
\begin{eqnarray}
\tilde{\epsilon}_\theta \left(x_t, t, c \right) = \left(1+w \right) \epsilon_\theta \left(x_t, c \right) - w \epsilon_\theta \left(x_t, \emptyset \right), \label{eq:score}
\end{eqnarray}
with the conditioning strength $w$. We set $w = 3$ for sampling. We employ the 1D U-Net \cite{RonnebergerMICCAI2015} based model to approximate $\epsilon_\theta \left(x_t, c \right)$. In the training, we use a single model to parameterize the conditional $\epsilon_\theta \left(x_t, c \right)$ and unconditional $\epsilon_\theta \left(x_t, \emptyset \right)$ scores by randomly discarding the conditioning during training \cite{HoNeurIPS2021}. For each observation interval, we train a separate diffusion model with $T=1000$. The noisy and sparse ``Observations'' $c$ are given in the same shape as the original data with zero filling wherein ``Observations'' are unavailable. \redtext{It should be noted that the unconditional labels are given by learnable parameters which are different from zeros in observations.} To accelerate the sampling procedure, we use the Denoising Diffusion Implicit Model (DDIM) \cite{songICLR2021} \redtext{with $T = 100$}, rather than directly using DDPM. We train the model for $100000$ steps with the batch size of 16. We use the Adam optimizer \cite{KingmaICLR2015} with the learning rate $0.001$.

% For DDIM, we set $T = 100$ and 

\subsection{DA method}
\label{eq:da_method}
Figure \figref{architecture} illustrates the DA process using DDPM. In each DA step, we generate pseudo ensembles with pre-trained diffusion models from noises guided by ``Observation'' data. We apply LETKF to these pseudo ensembles to compute analysis state vectors. We consider the ensemble mean of state vectors as the best estimate. The average of the ensemble mean and the simulation \redtext{state vector} is considered as the data-assimilated simulation \redtext{state}. 
\redtext{In EnKF, the numerical model is used to forecast the next states and the observations are used to alter the current states of the model. This way, the numerical models can start prediction from better model states. In the proposed method, the numerical model is used to generate the dataset and DDPM is trained to reproduce its data distribution guided by the observations. The generated ensembles are thus close to the observations. In our method, the information from past observation is carried through the data-assimilated simulation state vector which is the average of the ensemble and the simulation state vector.}
%\redtext{Table summarizes the comparison with conventional EnKF and our method which may be helpful to understand the proposed method.}

%The average of the ensemble mean and the simulation data is considered as the data-assimilated simulation data.

%to compute the ensemble mean. The average of the ensemble mean and the simulation data is considered as the data-assimilated simulation data.
%We compare the performance of our model with the well established DA method the local ensemble transform Kalman filter (LETKF) 
%For each DA step, we generate pseudo ensembles with pre-trained diffusion models from the observation data. To accelerate the sampling procedure, we use the Denoising Diffusion Implicit Model (DDIM) \cite{songICLR2021}. 

%In each diffusion step $t \left(t = 1, \ldots, T \right)$, a Gaussian noise $q \left(x_{t}|x_{t-1} \right):= {\cal N} \left(\sqrt{1-\beta_t} x_{t-1}, \beta_t \mathbf{I} \right)$ 
%In short, we define our DA method in the context of a neural style transfer. The simulation state is transferred into ensembles of mock simulation states where the style guidance is by mock observation data. 

%Figure \figref{architecture} illustrates the DA process using the diffusion model. In each data assimilation step, we apply forward DDIM processes to the simulation state followed by reverse DDIM processes. By setting $\sigma > 0$, the forward process can be stochastic and thus we can generate pseudo ensembles. We compare the performance of our model with the well established DA method the local ensemble transform Kalman filter (LETKF) \cite{HuntPhysicaD2007}. 

\begin{figure}[!t]
\vskip 0.2in
\begin{center}
\centerline{\includegraphics[clip, width=1.0\hsize]{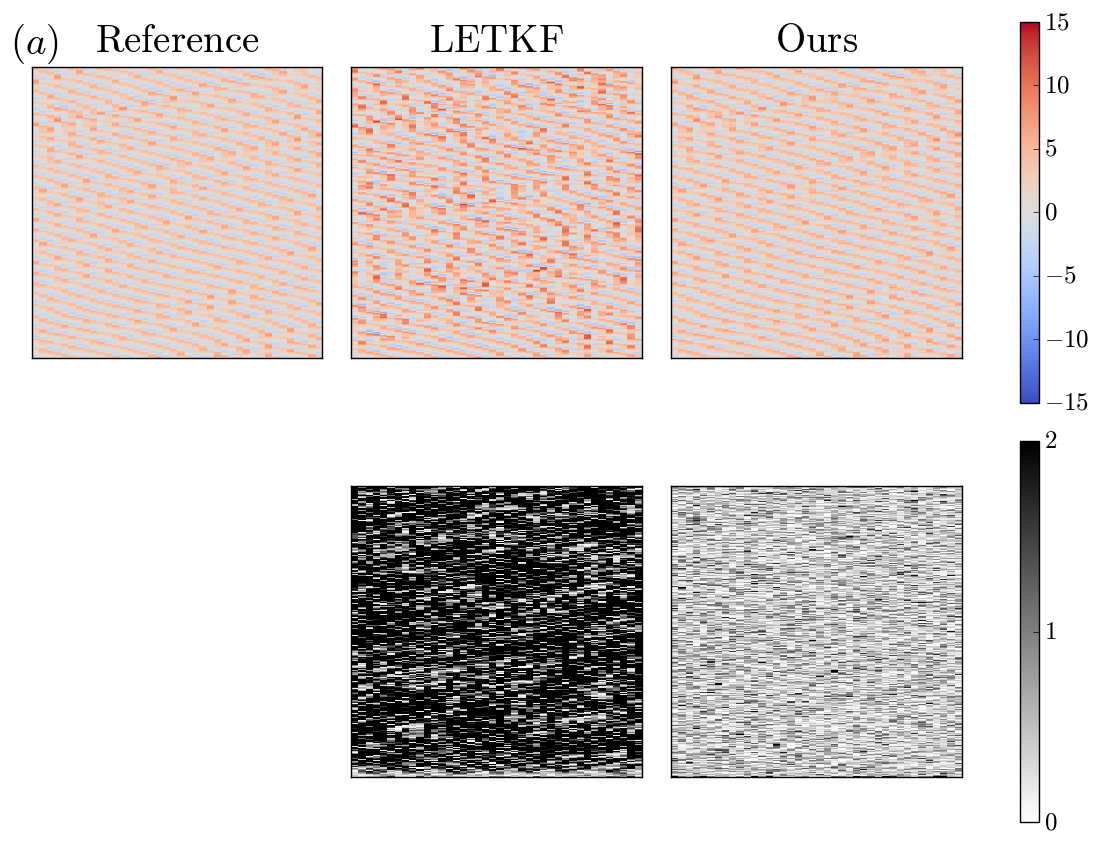}}
\centerline{\includegraphics[clip, width=1.0\hsize]{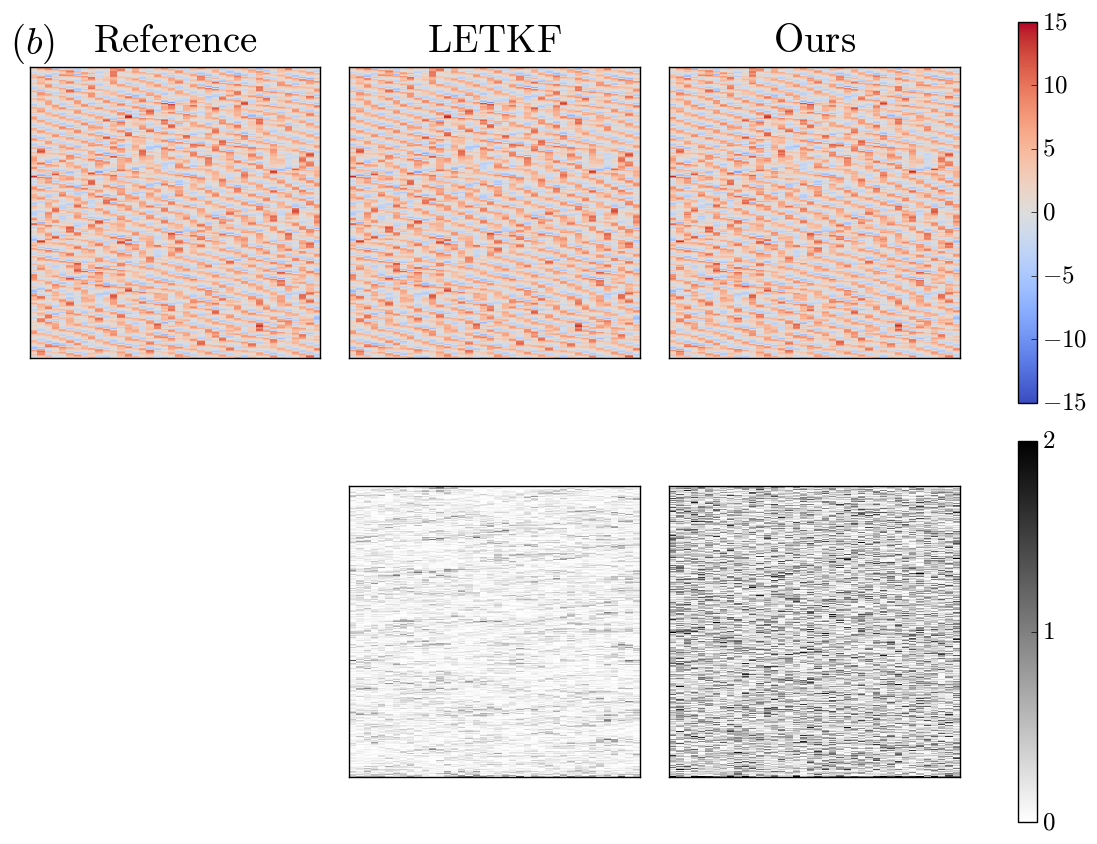}}
\centerline{\includegraphics[clip, width=1.0\hsize]{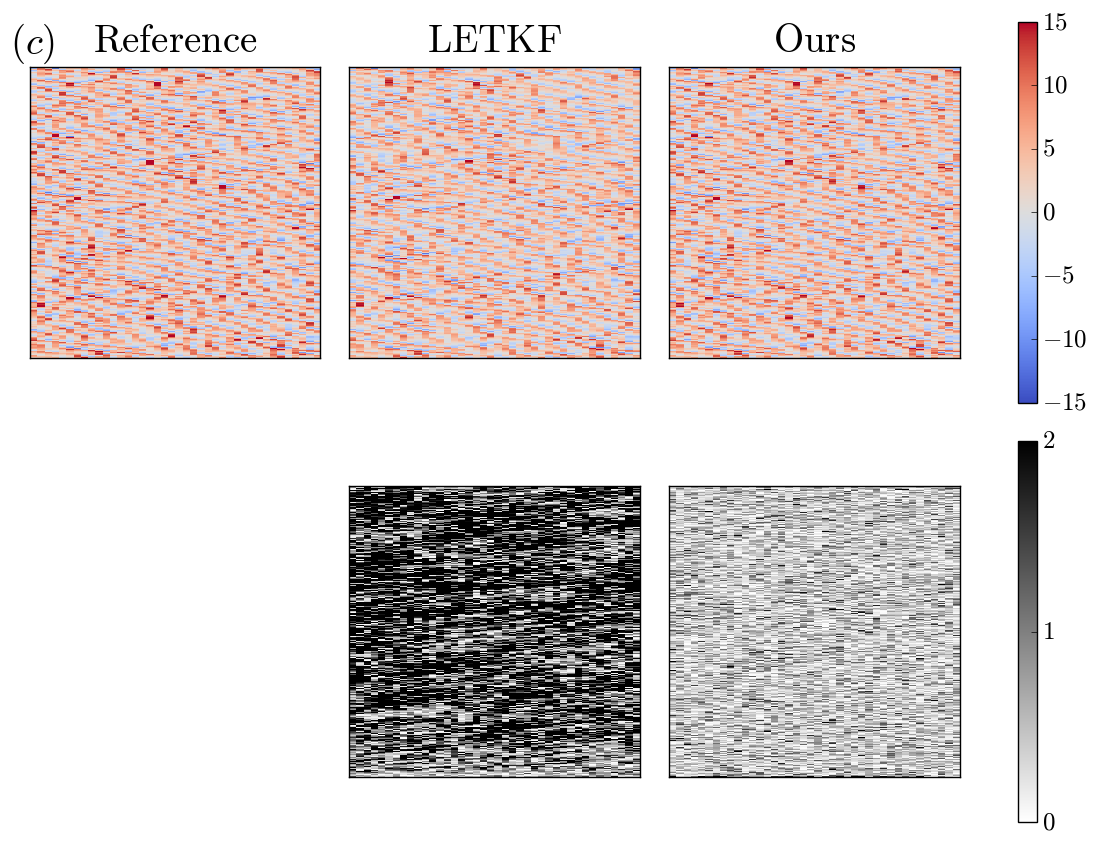}}
\caption{Hovm\"{o}ller diagrams and errors with (a) $F=5$, (b) $F=8$, and (c) $F=11$ cases. The upper row includes the Hovm\"{o}ller diagrams of reference, LETKF and our method, respectively. The bottom row includes the absolute errors between reference and DA simulations with LETKF and our method.}
\label{fig:spatio_temporal}
\end{center}
\vskip -0.2in
\end{figure}

\begin{figure}[t]
\vskip 0.2in
\begin{center}
\centerline{\includegraphics[clip, width=1.0\hsize]{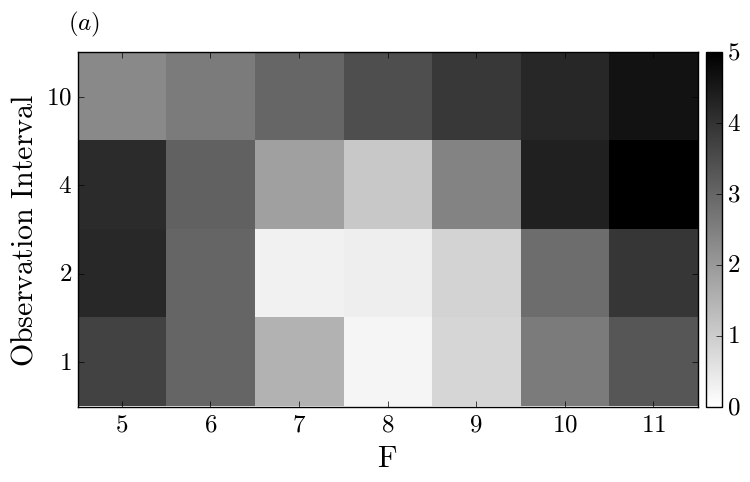}}
\centerline{\includegraphics[clip, width=1.0\hsize]{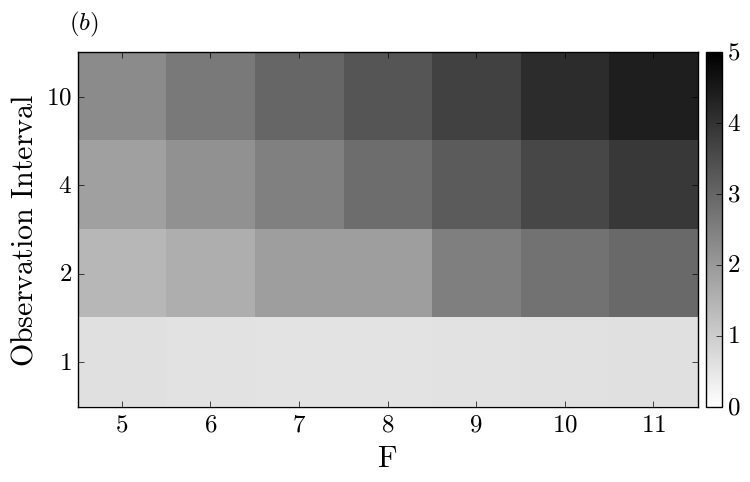}}
\caption{The dependency of RMSE with respect to the model bias and the observation interval. (a) LETKF and (b) our method.}
\label{fig:rmse}
\end{center}
\vskip -0.2in
\end{figure}

\section{Experiment}
\label{sec:experiment}
In this section, we compare the performance of DA with LETKF and ours.
We trained DDPM guided by sparse and noisy ``Observation'' data for the simulations with $F=8$. For LETKF, we perform 32 ensemble simulations. For our method, we perform a \redtext{non-ensemble} simulation and generate 32 ensembles from observation data. We compute
the analysis state vectors by applying LETKF to these ensembles. Then, we update the simulation state by the average of the ensemble mean and the current simulation state. DA is applied at each time step. We perform simulations with $F=8$ and assimilate the simulation states with the ``Observation'' data obtained from different ``Nature'' runs using multiple $F$ values with $F \in \left[5-11 \right]$. \redtext{Although we expect a performance gain with the proposed method for more complicated numerical models, DDIM sampling to generate ensembles is more costly than ensemble simulations of Lorenz96 system. The performance aspect of the proposed method will be discussed with more complicated applications in a separate publication.}

Figure \figref{spatio_temporal} shows the Hovm\"{o}ller diagrams of ``Nature'' runs and simulations with DA for the observation interval of 1 (40 observation points). %simulations with DA for the observation interval of 2 (20 observation points). 
As found in Fig. \figref{spatio_temporal} (b), when the model is not biased (the $F$ value is the same or similar for simulation and observation), LETKF displays better accuracy than ours. For $F=5$ and $F=11$, our method exhibits better accuracy than LETKF. In these biased cases, LETKF fails due to the filter divergence issue \cite{NgTellus2011}. The filter becomes overconfident around an incorrect state and thus the subsequent observations are ignored. It happens when the variance of ensembles is too small. The estimate cannot be moved back toward the true state. 
In contrast, our model always tracks the observations and the variance of ensembles is kept due to the properties of DDPM. Accordingly, our model stays closer to ``Observation'' data and filter divergence can be suppressed. %It is also shown that 

We investigate the impact of model biases (imperfectness of model) and observation intervals (imperfectness of observation) on the performance of DA methods. Figure \figref{rmse} shows the time averaged root mean square errors (RMSEs) between reference and DA simulations with LETKF and our method. As shown in Fig. \figref{rmse} (a), LETKF gives good prediction up to observation interval of 4 if ``Nature'' runs are made with $F \in \left[7-9 \right]$. Other than these cases, our model exhibits better performance as shown in Fig. \figref{rmse} (b). %our model in turn shows better performance except for these cases as shown in Fig. \figref{rmse} (b).
Our model outperforms LETKF when the incompleteness of models and observations are large. This is a preferable behaviour for DA considering that model biases are basically inevitable in real life.
%In Fig. \figref{spatio_temporal} (a),
%As found, when the $F$ value is the same for simulation and observation, LETKF displays better accuracy than ours. In contrast, our model 

\section{Summary}
\label{sec:summary}
In this work, we have proposed a DA method that relies on pseudo ensembles generated by DDPM. It has turned out that the proposed DA method is robust against the imperfectness of models (model biases) and observations (noises and sparsity). By using DDPM guided by noisy and sparse observations, we can generate divergent pseudo ensembles close to observation data. We can update the simulation result in the framework of LETKF using these ensembles. We compare the prediction accuracy of proposed DA method against LETKF with ensemble simulations. The proposed method outperforms LETKF particularly when the model is biased.
%when model bias is present.
%In this work, we have proposed a DA method without performing ensemble simulations. It has turned out that the proposed DA method is robust against the imperfectness of models (model biases) and observations (noises and sparsity). By using the DDPM guided by noisy and sparse observations, we can generate pseudo ensembles while keeping ensemble diversity. We can update the simulation result in the framework of LETKF using these ensembles. We compare the prediction accuracy of proposed DA method against LETKF with ensemble simulations. The proposed method outperforms LETKF when model bias is present.

\section*{Accessibility}
For better color accessibility, we use both multiple colors and line types in Figure 1. In order to display the error levels in Figs. 2 and 3, we avoid using color and show the intensity in gray scale. For all figures, we have enlarged the captions. %The contour plot is given in blue and red colors in Fig. 2.

%Authors are kindly asked to make their submissions as accessible as possible for everyone including people with disabilities and sensory or neurological differences.
%Tips of how to achieve this and what to pay attention to will be provided on the conference website \url{http://icml.cc/}.

\section*{Software and Data}
The source codes and dataset are available in the Github repository \cite{GenerativeEnKF}.

% Please do not forget to include a broader impact section!
\section*{Broader impact}
We have demonstrated the performance of NN based DA method in a simple 1D chaotic system. The proposed method is robust against the imperfectness of models (model biases) and observations (noises and sparsity). In principle, this method is applicable to 2D or even 3D simulations, which are used in various scientific domains.

%We have built a foundation to learn PDEs in a steerable manner rather than focusing on a specific application. Because of that, we envisage minimal risk of direct abusing our present work. However, as mentioned in the introduction, solving PDEs has many impacts on various domains, from both positive and negative aspects. Thus, our work and possible successive ones may be abused, aiming to harm lives and the environment. Therefore, the research community, including us, must be careful in using them and control the research direction to prevent abusing these technologies.

% Acknowledgements should only appear in the accepted version.
\section*{Acknowledgements}
This work was carried out using Tsubame 3.0 supercomputer at Tokyo Tech, HPE SGI8600 at JAEA, and FUJITSU PRIMERGY GX2570 (Wisteria/BDEC-01) at The University of Tokyo. This work was partly supported by JHPCN projects jh220031 and jh230033. This work has also received funding from JSPS KAKENHI Grant Number 23K11129.

% In the unusual situation where you want a paper to appear in the
% references without citing it in the main text, use \nocite
\nocite{langley00}

\bibliographystyle{synsml2023}

%%%%%%%%%%%%%%%%%%%%%%%%%%%%%%%%%%%%%%%%%%%%%%%%%%%%%%%%%%%%%%%%%%%%%%%%%%%%%%%
%%%%%%%%%%%%%%%%%%%%%%%%%%%%%%%%%%%%%%%%%%%%%%%%%%%%%%%%%%%%%%%%%%%%%%%%%%%%%%%
%%%%%%%%%%%%%%%%%%%%%%%%%%%%%%%%%%%%%%%%%%%%%%%%%%%%%%%%%%%%%%%%%%%%%%%%%%%%%%%
%%%%%%%%%%%%%%%%%%%%%%%%%%%%%%%%%%%%%%%%%%%%%%%%%%%%%%%%%%%%%%%%%%%%%%%%%%%%%%%

\clearpage
\newpage
\appendix
\section*{Appendix}
\label{sec:appendix}
%Although the Local ensemble transform Kalman Filter (LETKF) is not the main focus of this paper, it may be worth explaining how it works.   which is a kind of ensemble Data assimilation (DA) method.
%Here, we describe the details of Local ensemble transform Kalman Filter (LETKF) \cite{HuntPhysicaD2007} in the context of ensemble Data assimilation (DA). In general, DA deals with a dynamical system defined by
%\begin{eqnarray}
%\vect{x}^f_j &=& \mathbf{M}_j \vect{x}^a_{j-1} + \vect{\eta}_j,\\
%\vect{y}_j^o &=& \mathbf{H}_j \vect{x}^f_{j} + \vect{\epsilon}_j,
%\end{eqnarray}
%where $\vect{x}^f_j$ denotes the forecast variables, $\vect{x}^a_{j-1}$ denotes the analysis variables, and $\vect{y}^o_j$ is a vector of observed values. ${\cal M}_j$ represents the nonlinear dynamical model for time evolution from time $t_{j-1}$ to time $t_{j}$ with the stochastic model error $\vect{\eta}_j$. ${\cal H}_j$ is the observation operator which projects the forecast variables into the observation space with the observation error $\vect{\epsilon}_j$. These errors are assumed to be non-biased Gaussian distributions.

%\begin{eqnarray}
%\vect{x}^f_j &=& {\cal M}_j \left(\vect{x}^a_{j-1} \right) + \vect{\eta}_j,\\
%\vect{y}_j^o &=& {\cal H}_j \left(\vect{x}^f_{j} \right) + \vect{\epsilon}_j,
%\end{eqnarray}

Here, we describe the details of Local ensemble transform Kalman Filter (LETKF) \cite{HuntPhysicaD2007} in the context of ensemble Data assimilation (DA). In general, DA deals with a dynamical system defined by
\begin{eqnarray}
\vect{x}^f_j &=& {\cal M}_j \left(\vect{x}^a_{j-1} \right),\\
\vect{y}_j^o &=& {\cal H}_j \left(\vect{x}^t_{j} \right) + \vect{\epsilon}_j,~\vect{\epsilon}_j\sim {\cal N} \left(0, \mathbf{R}_j \right), \\
\vect{x}^a_{j} &=& \vect{x}^t_{j} + \vect{\eta}_j,~\vect{\eta}_j\sim {\cal N} \left(0, \mathbf{P}_j^a \right)
\end{eqnarray}
where $\vect{x}^f_j$ denotes the forecast variables, $\vect{x}^a_{j-1}$ denotes the analysis variables, \redtext{$\vect{x}^t_j$ represents the ``true'' state, and $\vect{y}^o_j$ is a vector of observed values. ${\cal M}_j$ represents the nonlinear dynamical model for time evolution from time $t_{j-1}$ to time $t_{j}$. ${\cal H}_j$ is the observation operator which projects the forecast variables into the observation space with the observation error $\vect{\epsilon}_j$. The analysis variables $\vect{x}^a_{j}$ are assumed to be apart from the ``true'' state $\vect{x}^t_{j}$ with the stochastic error $\vect{\eta}_j$.
These errors are respectively assumed to be non-biased (zero mean) Gaussian distributions with the covariance matrices $\mathbf{R}_j$ and $\mathbf{P}_j^a$.}
%with the stochastic forecast error $\vect{\eta}_j$. ${\cal H}_j$ is the observation operator which projects the forecast variables into the observation space with the observation error $\vect{\epsilon}_j$. These errors are assumed to be non-biased Gaussian distributions.

In DA, the model is used to forecast variables at the next time step from the current analysis variables (forecast step). Then, the current observation data are used to update the prior forecast variables to current state estimates which maximize the Bayesian likelihood (analysis step).

%simulations with DA consist of forecast and analysis steps. 

%In the forecast step, the time integration of the numerical model is computed to get the forecast state vector. In the analysis step, the forecast state vector is altered based on the observation vector.

LETKF is based on Kalman Filter (KF) which estimates the state vector $\vect{x}^a_j$ at time $t_j$ with the following \redtext{linearized} equations.
\begin{eqnarray}
\vect{x}^a_j &=& \vect{x}^f_j + \mathbf{K}_j \left(\vect{y}_j^o - \mathbf{H}_j \vect{x}_j^f \right), \\
\mathbf{K}_j &=& \mathbf{P}_j^f \mathbf{H}_j^\top \left(\mathbf{H}_j \mathbf{P}_j^f \mathbf{H}_j^\top + \mathbf{R}_j\right)^{-1}, \label{eq:Kalman}\\
\mathbf{P}_j^a &=& \left(\mathbf{I} - \mathbf{K}_j \mathbf{H}_j\right) \mathbf{P}_j^f,
\end{eqnarray}
where the superscripts $f$ and $a$ respectively denote the forecast and analysis, $\vect{y}_j^o$ is the observation vector, $\mathbf{I}$ is the identity matrix, $\mathbf{H}_j$ is the observation matrix which projects the state vector $\vect{x}_j^f$ into the observation space, and $\mathbf{K}_j$ is a matrix called the Kalman gain. \redtext{Here, matrix $\mathbf{H}_j$ is the linearization of the observation operator ${\cal H}_j$. The time evolution operator is also linearlized to give $\vect{x}^f_j = \mathbf{M}_j \vect{x}^a_{j-1}$. The Kalman gain $\mathbf{K}_j$ defined by eq. (\eqnref{Kalman}) gives the most likely state $\vect{x}^a_j$ given the observations up to time $t_j$ in a least square sense. For simplicity, we hereinafter omit the subscript $j$ representing time.}

%multiplies the difference between the observations and the } 
%If the Kalman gain is given by eq. (\eqnref{Kalman}), the analysis error between true state vector and analysis state vector ($\vect{x}^t - \vect{x}^a$) is minimized, where $\vect{x}^t$ represents the true state vector $\vect{x}^t$.

Unfortunately, when the number of model variables $m$ is huge (this is often the case for real-world models such as a global weather model), computing the covariance matrix $\mathbf{P}$ is a formidable task involving the $m \times m$ matrix operations. Instead, we use ensembles of model states to approximate $\mathbf{P}$. 

In LETKF, the analysis vector $\vect{x}^a$ of the $m$-th ensemble is computed by
\begin{eqnarray}
\vect{x}_m^a = \overline{\vect{x}}^f + \delta \mathbf{X}^f \left( \overline{\vect{w}}^a + \delta \vect{W}_m^a \right),
\end{eqnarray}
where $\overline{\vect{w}}^a$ and $\delta \vect{W}_m^a$ are the mean of the transformed vector and the $m$-th column of the perturbation of the transformed vector. These are calculated by
\begin{eqnarray}
\overline{\vect{w}}^a &=& \tilde{\mathbf{P}}^a \delta \mathbf{Y}^{f\top} \mathbf{R}^{-1} \left(\vect{y}^o - \overline{\mathbf{y}}^f\right), \\
\delta \vect{W}^a &=&  \left[ \left(N_e-1 \right)
\tilde{\mathbf{P}}^a\right]^{1/2},
\end{eqnarray}
where $N_e$ is the number of ensembles. $\vect{y}^f \left(= \mathbf{H} \vect{x}^f\right)$ is the observation vector computed from the model state $\vect{x}^f$. The inflation coefficient $\beta$ is introduced to mitigate the so-called filter divergence problem, resulting from the small error covariance. The matrix $\mathbf{R}$ represents the observation error covariance, on which localization is applied to suppress the spurious long-distance correlation.
The covariance matrix $\tilde{\mathbf{P}}^a$ is defined as
\begin{eqnarray}
\tilde{\mathbf{P}}^a = \left[\frac{\left(N_e-1 \right)}{\beta} \mathbf{I} + \delta \mathbf{Y}^{f\top} \mathbf{R}^{-1} \delta \mathbf{Y}^{f} \right]^{-1} \equiv \mathbf{Q}^{-1}.
\end{eqnarray}
The ensemble mean $\overline{\vect{x}}^f$ and $\overline{\vect{y}}^f$, and perturbation matrices $\delta \mathbf{X}^f$ and $\delta \mathbf{Y}^f$ are defined by
\begin{eqnarray}
\overline{\vect{x}}^f &=& \frac{1}{N_e} \sum_{m=0}^{N_e-1} \vect{x}_m^f, \\
\delta \mathbf{X}^f &=& \left[\vect{x}_0^f - \overline{\vect{x}}^f, \vect{x}_1^f - \overline{\vect{x}}^f, \ldots, \vect{x}_{N_e-1}^f - \overline{\vect{x}}^f \right], \\
\overline{\vect{y}}^f &=& \frac{1}{N_e} \sum_{m=0}^{N_e-1} \vect{y}_m^f, \\
\delta \mathbf{Y}^f &=& \left[\vect{y}_0^f - \overline{\vect{y}}^f, \vect{y}_1^f - \overline{\vect{y}}^f, \ldots, \vect{y}_{N_e-1}^f - \overline{\vect{y}}^f \right].
\end{eqnarray}

$\tilde{\mathbf{P}}^a$ and $\left[\tilde{\mathbf{P}}^a\right]^{1/2}$ can be efficiently computed by the Eigen Value Decomposition (EVD) of $\mathbf{Q}$.
In LETKF, we apply the localization on $\mathbf{R}$ as $\mathbf{R}_{\rm loc}^{-1} = \mathbf{G} \circ \mathbf{R}^{-1}$, which decompose the analysis procedure into a local problem on each grid point. As the local problem is independent on each grid point, LETKF is highly parallelizable and thus suitable for many recent parallel architectures such as GPUs. 
In the present work, the localization function is given by the Gaspari-Cohn function \cite{GaspariRMETS1999}, in which the cutoff distance is chosen as $d = 2 \left(p-1 \right) \Delta x$ based on the distance between observation points $p\Delta x$, where $p$ and $\Delta x$ denote the observation interval and grid width, respectively.

\end{document}